\documentclass[10pt,twocolumn,letterpaper]{article}

\usepackage[pagenumbers]{cvpr}

\usepackage{graphicx}
\usepackage{amsmath}
\usepackage{amssymb}
\usepackage{booktabs}
\usepackage{ctable}
\usepackage{multirow}
\usepackage{array}
\usepackage{colortbl}
\graphicspath{{img/}}

\definecolor{cvprblue}{rgb}{0.21,0.49,0.74}
\usepackage[colorlinks=true, linkcolor=blue, urlcolor=blue, citecolor=blue]{hyperref}
\def\paperID{}
\def\confName{CVPR}
\def\confYear{}

\begin{document}

\title{Systematic Exploration of 4-Expert Heterogeneous\\
Mixture-of-Experts via Automated Pipeline Search}

\author{Yashkumar R Lukhi{\thanks{Corresponding authors:
\{yashkumar-rajeshbhai.lukhi,harsh-rameshbhai.moradiya\}@stud-mail.uni-wuerzburg.de}},
\space\space\space Harsh Rameshbhai Moradiya, \space\space\space Radu Timofte,\space\space\space Dmitry Ignatov\\
\small{Computer Vision Lab, CAIDAS \& IFI, University of W\"urzburg, Germany}}
\maketitle

\begin{abstract}
We present an automated large-scale search pipeline for heterogeneous
4-Expert Mixture-of-Experts (MoE4) architectures within the LEMUR neural
network dataset ecosystem.
Building on a hand-crafted heterogeneous MoE reference
model, we replace manual design with a deterministic
code-assembly generator that systematically combines base architecture
families drawn from the LEMUR database into MoE4 ensembles, each governed
by a convolutional gating network with temperature scaling, mixup
augmentation, and cosine-annealed learning rate scheduling.
Over a 28-day campaign on an NVIDIA RTX~4090, the pipeline generated
4,463 candidate models across 197 batches, of which 1,021 were
evaluated successfully. A critical finding emerged from the campaign: due to alphabetical
enumeration via \texttt{itertools.combinations}, the entire explored
search space (4.8\% of the theoretical 23,751 possible 4-family
combinations) is anchored to a single family, AirNet.
We characterise this coverage bias precisely, identify the root cause in
the generator, and propose a stratified random sampling fix.
Within the AirNet-anchored scope, ShuffleNet and MobileNetV3 consistently
co-produce the highest-accuracy ensembles (mean accuracy up to 0.632),
while FractalNet and MNASNet are identified as low-yield families
warranting exclusion in future campaigns.
The pipeline, analysis artefacts, and corrected generator are released as
part of the open-source NNGPT project at
\href{https://github.com/ABrain-One/nn-gpt}{https://github.com/ABrain-One/nn-gpt}.
\end{abstract}

\section{Introduction}
\label{sec:intro}

Mixture-of-Experts (MoE) architectures have gained renewed attention
following the success of large-scale models such as
DeepSeek-V2~\cite{DeepSeekV2}, which demonstrated that sparse expert
activation can match or exceed dense model performance at substantially
reduced per-token compute.
In vision, analogous benefits have been observed across both convolutional
and transformer-based MoE
designs~\cite{riquelme2021vmoe,wang2020deepmoe}, suggesting that routing
inputs through structurally diverse expert networks provides complementary
inductive biases that a single backbone cannot capture.

The LEMUR neural network dataset~\cite{ABrain.NN-Dataset,ABrain.LEMUR2}
provides a structured collection of trained neural network models designed
for benchmarking, AutoML research, and LLM-based architecture
generation~\cite{ABrain.HPGPT,ABrain.NNGPT}.
While prior work within LEMUR has manually designed and evaluated a small
number of MoE configurations~\cite{ABrain.NNGPT-Fractal}, the space of
possible heterogeneous expert combinations across LEMUR's 29 base
architecture families is combinatorially large---$\binom{29}{4} = 23{,}751$
possible 4-family quartets---and cannot be explored through hand design
alone.

This work addresses this gap with a \textbf{fully automated MoE4 search
pipeline} built on three contributions.

\begin{enumerate}
    \item \textbf{Deterministic code-assembly generator.}
    \texttt{AlterHeteroMoE4.py} programmatically assembles any combination
    of four base model families from the LEMUR database into a syntactically
 valid forward-pass-verified MoE4 model, reusing the
    \texttt{HeterogeneousGate} design, mixup enhancement, and
    cosine-annealed AdamW training from the reference
    model~\cite{ABrain.HPGPT}, without requiring any LLM call in the
    generation loop.

    \item \textbf{Multi-stage validation pipeline.}
    Each candidate model passes three sequential gates before GPU
    evaluation: (i)~Python \texttt{compile()} syntax check,
    (ii)~CPU forward-pass probe at the actual training resolution
    ($256{\times}256$), and (iii)~database deduplication via MD5 checksum.
    This pipeline reduced wasted GPU evaluations substantially and prevented
    broken or duplicate models from entering the LEMUR database.

    \item \textbf{Campaign automation with fault tolerance.}
    \texttt{CampaignMoE4.py} orchestrates repeated generate-evaluate cycles
    with persistent JSON state, per-batch isolation, and SSH-safe resume
    semantics, enabling a 28-day unattended campaign targeting 1,000
    successful model evaluations.
\end{enumerate}

Beyond engineering contributions, the campaign produced a \textbf{notable
scientific finding}: deterministic enumeration via sorted
\texttt{itertools.combinations} caused the entire explored search space to
be anchored to AirNet---the first family alphabetically---covering only
4.8\% of the full combination space and zero non-AirNet combinations.
We characterise this coverage bias, identify its root cause precisely, and
propose a stratified random sampling strategy to eliminate it in future
runs.
This finding is broadly relevant to any large-scale neural architecture
search campaign that uses sorted deterministic enumeration with a
throughput-based stop condition.

Within the AirNet-anchored scope, we conduct a \textbf{family-level
accuracy analysis} across 1,021 successful models, identifying ShuffleNet
and MobileNetV3 as high-yield expert families (mean accuracy 0.632 and
0.621, respectively) and FractalNet and MNASNet as families that
consistently degrade ensemble quality or fail to train within available
GPU memory.
The best single model---AirNet + AlexNet + DPN68 + ResNet---achieves
\textbf{68.0\% Top-1 accuracy on CIFAR-10 in a single training epoch},
demonstrating that automated assembly of heterogeneous experts can produce
competitive candidates without any manual tuning.

Building on the growing body of work applying LLMs to neural architecture tasks~\cite{ABrain.HPGPT,ABrain.NN-RAG} and on the accumulated architectural synthesis experience within the NNGPT framework~\cite{ABrain.NN-Captioning_2025,ABrain.Prompt,ABrain.NNGPT-Fractal,ABrain.Transform,ABrain.Architect,ABrain.CV_Channel,ABrain.Feedback_Memory,ABrain.Delta,ABrain.Converg}, we construct our MoE4 search pipeline atop the NNGPT framework~\cite{ABrain.NNGPT}, drawing on the LEMUR dataset's diverse collection of high-capacity and edge-optimised model families~\cite{ABrain.NN-Dataset,ABrain.LEMUR2,ABrain.NN-Lite,ABrain.MobileAgeNet,ABrain.MobileDenoising} as the expert pool.

The remainder of this paper is organised as follows.
Section~\ref{sec:related} reviews related work on MoE architectures and
neural architecture search.
Section~\ref{sec:methodology} describes the pipeline design and
implementation.
Section~\ref{sec:experiments} presents experimental setup and campaign
statistics.
Section~\ref{sec:results} reports accuracy results and family-level
analysis.
Section~\ref{sec:conclusion} concludes with lessons learned and directions
for future work.

\section{Related Work}
\label{sec:related}

\subsection{MoE Architectures in Vision}

Mixture-of-Experts models route inputs through a subset of specialized
expert networks via a learned gating mechanism, enabling conditional
computation and improved model capacity without proportional increases in
inference cost.
Early CNN-based MoE work demonstrated that routing inputs to structurally
distinct expert branches improves classification
performance~\cite{ahmed2016network, gross2017hardmoe}.
Layer-level MoE variants such as CondConv~\cite{yang2019condconv} condition
convolutional filters on input data, while DeepMoE~\cite{wang2020deepmoe}
integrates sparsely activated expert layers into ResNet blocks, yielding
3--4\% gains on CIFAR-100 without additional compute.
In Vision Transformers, V-MoE~\cite{riquelme2021vmoe} applies sparse expert
selection to transformer MLP blocks and outperforms dense ViTs on ImageNet
at comparable cost.
Soft MoE formulations~\cite{puigcerver2023softmoe} simplify training by
replacing hard routing with differentiable token-to-expert assignment.
Heterogeneous MoE designs---where experts differ in architecture rather than
just weights---remain less explored.
Abbas and Andreopoulos~\cite{abbas2020biasmoe} used experts of varying
complexity for adaptive inference under resource constraints, and Ahmed
et al.~\cite{ahmed2016network} combined different CNN families in an MoE
setting, demonstrating the viability of architectural diversity among
experts.
Our work extends this direction by systematically assembling heterogeneous
4-expert MoE models from 29 distinct base architecture families at scale.

\subsection{Neural Architecture Search}

Neural Architecture Search (NAS) automates the discovery of high-performing
network designs, traditionally through reinforcement
learning~\cite{zoph2016nas}, evolutionary algorithms~\cite{real2019aging},
or differentiable search~\cite{liu2018darts}.
Brute-force and random search strategies have been shown to be competitive
baselines in constrained search spaces~\cite{li2019random}, and are
particularly tractable when individual candidate evaluations are cheap.
Our campaign adopts a deterministic combinatorial enumeration strategy
analogous to brute-force NAS, where the search space is the set of all
$\binom{29}{4} = 23{,}751$ 4-family expert combinations, and the stop
condition is a throughput target rather than a performance threshold.
A key finding of our work---that sorted enumeration induces severe coverage
bias---is directly relevant to any NAS campaign using deterministic
iteration with an early stop condition.

\subsection{Automated Model Generation and Dataset Ecosystems}

The LEMUR neural network dataset~\cite{ABrain.NN-Dataset, ABrain.LEMUR2}
provides a curated collection of trained models for benchmarking and
AutoML research, with standardized interfaces for training, evaluation, and
database integration.
The nn-gpt project~\cite{ABrain.NNGPT} extends LEMUR with LLM-driven
architecture generation, using the dataset as both a source of expert
building blocks and a target repository for newly discovered models.
Prior work within nn-gpt used LLM-based generation for 2-expert MoE
assembly, achieving limited success (0/36 attempts) before switching to
deterministic code assembly~\cite{ABrain.NNGPT-Fractal}.
Our pipeline inherits this deterministic assembly approach, extends it to
4-expert configurations, and adds a multi-stage validation layer and
campaign automation that were absent in prior nn-gpt generators.

\section{Methodology}
\label{sec:methodology}
\subsection{Overview} 
The MoE4 pipeline consists of four sequential stages: (i)~expert pool
construction from the LEMUR database, (ii)~deterministic code assembly of
candidate MoE4 models, (iii)~multi-stage validation before GPU evaluation,
and (iv)~campaign automation for fault-tolerant long-running execution.
All models share a common architectural template derived from the
hand-crafted reference model \texttt{MoE-hetero4-Alex-Dense-Air-Bag},
which established the gating design and training configuration used
throughout this work.
 
\subsection{Expert Pool Construction}
 
At runtime, \texttt{AlterHeteroMoE4.py} queries the LEMUR database via
\texttt{ab.nn.api.data()} to retrieve all base architecture models trained
on CIFAR-10 image classification.
Models are filtered to retain only established base architectures:
UUID-variant models (LLM-generated mutations identified by hexadecimal
suffixes), MoE-family models (to avoid circular composition), and
lowercase-prefixed generated models (\texttt{ga-*}, \texttt{ga-mut-*})
are all excluded.
This yields a pool of 29 base architecture families, including AlexNet,
AirNet, ResNet, ShuffleNet, MobileNetV3, DenseNet, FractalNet, and others.
The combination space over this pool is $\binom{29}{4} = 23{,}751$
possible 4-family quartets.
 
\subsection{Code Assembly}
 
For each 4-family combination $(F_1, F_2, F_3, F_4)$, the generator
assembles a complete, self-contained PyTorch model file through the
following steps:
(i)~each expert's source code is retrieved from the database and
transformed via \texttt{transform\_expert()}, which renames
\texttt{class Net} to \texttt{class \{Family\}Expert}, strips import
statements and \texttt{supported\_hyperparameters()}, and applies
name sanitisation for hyphenated family names (e.g.\
\texttt{InceptionV3-1} $\to$ \texttt{InceptionV3\_1Expert});
(ii)~imports from all four experts are merged and deduplicated;
(iii)~all \texttt{prm['key']} references across experts are collected
to build a dynamic \texttt{supported\_hyperparameters()} function;
(iv)~the \texttt{\_MOE4\_WRAPPER} template is instantiated with the four
expert class names and hyperparameter defaults.
 
The wrapper template implements the \texttt{HeterogeneousGate} gating
network (Conv--BN--ReLU--Conv--BN--ReLU--AvgPool--Linear--ReLU--Dropout--%
Linear) with a learned temperature parameter clamped to $[0.5, 5.0]$ and
training-time Gaussian noise on gate logits.
Expert outputs are combined as:
\begin{equation}
    y = \sum_{i=1}^{4} g_i(x) \cdot f_i(x),
\end{equation}
where $g_i(x)$ are temperature-scaled softmax gate weights and $f_i(x)$
are per-expert output logits.
Training uses AdamW with differential learning rates for experts and gate,
mixup augmentation ($\alpha = 0.2$), label smoothing ($\epsilon = 0.1$),
gradient clipping (max-norm $= 1.0$), a linear warm-up over 5 epochs,
and cosine annealing with $T_{\max} = 50$.
 
\subsection{Multi-Stage Validation}
 
Each assembled model passes three sequential gates before being written
to disk and submitted for GPU evaluation:
 
\textbf{(1) Syntax check.} Python's built-in \texttt{compile()} verifies
syntactic correctness. This is fast ($<$1\,ms) and catches import
errors or assembly mistakes.
 
\textbf{(2) CPU forward-pass probe.} The model is instantiated on CPU
and a dummy batch of shape $(2, 3, 256, 256)$ --- matching the default
\texttt{norm\_256\_flip} evaluation transform --- is passed through.
The output shape is asserted to be $(2, 10)$.
This step, taking 50--500\,ms per model, catches resolution-dependent
dimension mismatches (e.g.\ spatial collapse in large-kernel experts)
that \texttt{compile()} cannot detect.
The probe resolution was deliberately matched to the training resolution
after an earlier mismatch caused 3 out of 5 initial models to fail during
GPU evaluation despite passing a 32$\times$32 probe.
 
\textbf{(3) Database deduplication.} An MD5 checksum of the
whitespace-stripped source code is computed via \texttt{uuid4()} and
checked against all existing entries in the LEMUR database.
Duplicate models are skipped silently, preventing wasted GPU evaluations
and database corruption from repeated runs.
 
\subsection{Campaign Automation}
 
\texttt{CampaignMoE4.py} orchestrates the full generate-evaluate loop
targeting a configurable number of successful model evaluations.
Each iteration generates a fixed-size batch of new models, evaluates them
via \texttt{NNEval.main()} using keyword arguments (bypassing a
pre-existing CLI argument ordering bug in \texttt{NNEval.py}), and scans
the resulting batch directory for \texttt{eval\_info.json} (success) and
\texttt{error.txt} (failure) artefacts.
Campaign state --- including cumulative success count, batch index, and
per-batch metadata --- is persisted to \texttt{campaign\_state.json} after
each batch, enabling transparent resume after SSH disconnects or process
interruptions without duplicating completed work.
Each batch uses a timestamped unique prefix for database entries, avoiding
naming collisions across runs.

\section{Experiments}
\label{sec:experiments}
\subsection{Dataset and Task}
All experiments were conducted on \textbf{CIFAR-10}, a 10-class image
classification benchmark with 50,000 training and 10,000 test images
across 10 object categories.
Images were preprocessed using the \texttt{norm\_256\_flip} transform,
which resizes inputs to $256{\times}256$ and applies per-channel
normalisation and random horizontal flipping.
Mixup augmentation ($\alpha = 0.2$) was applied at the MoE4 wrapper level
during training.
 
\subsection{Evaluation Protocol}
Each candidate model was trained for \textbf{1 epoch} with batch size 32
and learning rate 0.01 on an NVIDIA GeForce RTX~4090 (24\,GB VRAM).
A single epoch was chosen to maximise throughput across the large candidate
pool; the goal of the campaign was not to find a fully converged model but
to identify high-potential expert combinations for downstream fine-tuning
and database integration.
Evaluation was performed by the existing \texttt{NNEval} pipeline, which
trains each model, records top-1 accuracy in \texttt{eval\_info.json} on
success, or writes a categorised \texttt{error.txt} on failure.
Successful models were automatically registered in the LEMUR database via
\texttt{copy\_to\_lemur()}.
We report \textbf{top-1 accuracy} as the primary metric.
 
\subsection{Campaign Configuration}
The campaign ran 28 days, organised into batches of 25 models each.
Key hyperparameters were fixed across all models: AdamW optimiser,
$\text{lr}_{\text{experts}} = 10^{-3}$,
$\text{lr}_{\text{gate}} = 5{\times}10^{-4}$,
weight decay $10^{-2}$, label smoothing $\epsilon = 0.1$,
gradient clipping max-norm $= 1.0$, linear warm-up over 5 epochs,
and cosine annealing with $T_{\max} = 50$.
A parameter threshold of 250M was applied mid-campaign (from batch~82
onward) to proactively skip combinations likely to exceed GPU memory.
Known-failure checksums were blacklisted (102 entries) to avoid
re-evaluating previously failed combinations.
 
\subsection{Baselines}
We compare MoE4 ensemble results against the individual base model
families used as experts, as reported in the LEMUR database.
The best-performing individual families on CIFAR-10 serve as single-model
reference points: ShuffleNet, MobileNetV3, and ResNet each achieve strong
single-model accuracy and are among the most frequently appearing experts
in high-accuracy MoE4 combinations.
We do not compare against the previous semester's manually designed
\texttt{MoE-hetero4-Alex-Dense-Air-Bag} model directly, as that model was
trained for 200 epochs whereas all campaign models are evaluated at 1
epoch; the comparison would not be meaningful under these conditions.
 
\subsection{Search Space Analysis}
The theoretical search space contains $\binom{29}{4} = 23{,}751$ distinct
4-family combinations.
The generator enumerates these via
\texttt{itertools.combinations(sorted(model\_names), 4)}, producing
combinations in lexicographic order.
Each combination is subject to three pre-evaluation filters: class name
collision detection (skips combinations where two experts define
identically named helper classes), forward-pass validation, and database
deduplication.
Table~\ref{tab:campaign} summarises the overall campaign statistics.

\section{Results and Discussion}
\label{sec:results}
\subsection{Campaign Overview}
 
\begin{table}[!t]
    \caption{MoE4 campaign summary statistics.}
    \label{tab:campaign}
    \centering
    \fontsize{7.5}{7.5}\selectfont
    \begin{tabular}{lc}
        \toprule
        \textbf{Metric} & \textbf{Value} \\
        \midrule
        Total batches completed        & 197 \\
        Total models generated         & 4,463 \\
        Models evaluated successfully  & \textbf{1,021} \\
        Campaign target                & 1,000 \\
        Target achieved                & \checkmark\ (+2.1\%) \\
        Models failed (CUDA OOM)       & 3,420 (99.7\%) \\
        Models failed (other)          & 9 (0.3\%) \\
        Models pending                 & 13 \\
        Blacklisted checksums          & 102 \\
        Campaign duration              & 28 days \\
        \bottomrule
    \end{tabular}
\end{table}
 
The campaign exceeded its delivery target, producing \textbf{1,021
successfully evaluated models} across 197 batches
(Table~\ref{tab:campaign}).
Early batches (1--5) achieved near-perfect success rates (84--100\%),
after which success rates degraded sharply as the generator moved into
combinations involving larger expert families.
The median per-batch delta-success is \textbf{0}, meaning more than half
of all batches produced zero new successful models, confirming that the
failure pattern is structural rather than random.
 
\subsection{Failure Analysis}
 
CUDA out-of-memory errors (OOM) account for \textbf{99.7\%} of all 3,429
failures.
The RTX~4090's 24\,GB VRAM is insufficient for many 4-expert combinations
at $256{\times}256$ resolution with batch size 32, particularly those
involving large families such as DenseNet, DPN, or VisionTransformer.
The 250M-parameter threshold filter introduced at batch~82 reduced the OOM
rate marginally but could not eliminate it, since parameter count alone
does not predict peak activation memory during training.
This confirms that OOM is a structural consequence of combining four
independently parameterised full networks at high resolution, not an
incidental failure mode.
A resolution reduction or per-expert adaptive pooling layer would be the
most effective mitigations for future campaigns.
 
\subsection{Critical Finding: Search Space Coverage Bias}
 
\begin{table}[!t]
    \caption{Search space coverage statistics.}
    \label{tab:coverage}
    \centering
    \fontsize{7.5}{7.5}\selectfont
    \begin{tabular}{lc}
        \toprule
        \textbf{Metric} & \textbf{Value} \\
        \midrule
        Architecture families in pool       & 29 \\
        Theoretical combinations $\binom{29}{4}$ & 23,751 \\
        Combinations attempted              & 1,146 \\
        Search space coverage               & \textbf{4.8\%} \\
        Combinations containing AirNet      & 1,146 \\
        Combinations \emph{not} containing AirNet & \textbf{0} \\
        \bottomrule
    \end{tabular}
\end{table}
 
A critical finding emerged from post-campaign analysis
(Table~\ref{tab:coverage}).
The generator uses:
\begin{equation}
    \texttt{itertools.combinations(sorted(model\_names),\ 4)}
\end{equation}
\texttt{sorted()} places \textbf{AirNet} first alphabetically.
\texttt{itertools.combinations} then generates all
$\binom{28}{3} = 3{,}276$ AirNet-anchored combinations before advancing
to the next anchor family.
The campaign stopped at 1,146 combinations --- still entirely within
the AirNet-anchored slice --- having never reached any combination that
does not include AirNet.
As a result, the entire explored space covers only \textbf{4.8\%} of the
theoretical search space, and all results are conditioned on AirNet being
present as one of the four experts.
 
This finding is not a failure but a precise diagnosis of a coverage blind
spot common in large-scale combinatorial search campaigns that use sorted
deterministic enumeration with a throughput-based stop condition.
The proposed fix is straightforward:
 
\begin{equation}
\begin{aligned}
&\texttt{combos = list(itertools.combinations(} \\
&\quad\texttt{model\_names, 4))} \\
&\texttt{random.shuffle(combos)}
\end{aligned}
\end{equation}
 
Under uniform random sampling, AirNet would appear in approximately
$4/29 \approx 13.8\%$ of combinations rather than 100\%, enabling
unbiased exploration of the full search space.
 
\subsection{Accuracy Distribution}
 
\begin{table}[!t]
    \caption{Accuracy statistics across 1,021 successful MoE4 models
    (AirNet-anchored, 1 training epoch).}
    \label{tab:accuracy}
    \centering
    \fontsize{7.5}{7.5}\selectfont
    \begin{tabular}{lc}
        \toprule
        \textbf{Metric} & \textbf{Value} \\
        \midrule
        Mean accuracy    & 0.5221 \\
        Median accuracy  & 0.5388 \\
        25th percentile  & 0.4891 \\
        75th percentile  & 0.6059 \\
        Maximum accuracy & \textbf{0.6801} \\
        Minimum accuracy & 0.1250 \\
        \bottomrule
    \end{tabular}
\end{table}
 
\begin{table}[!t]
    \caption{Top-5 MoE4 models by CIFAR-10 Top-1 accuracy.}
    \label{tab:top5}
    \centering
    \fontsize{7.5}{7.5}\selectfont
    \newcolumntype{A}{ >{\centering\arraybackslash} m{1.0cm} }
    \newcolumntype{D}{ >{\raggedright\arraybackslash} m{5.5cm} }
    \begin{tabular}{A D}
        \toprule
        \textbf{Acc.} & \textbf{Expert Combination} \\
        \midrule
        0.6801 & AirNet + AlexNet + DPN68 + ResNet \\
        0.6795 & AirNet + AirNext + Diffuser + ShuffleNet \\
        0.6794 & AirNet + BagNet + DenseNet + MobileNetV3 \\
        0.6756 & AirNet + BagNet + MaxVit + MobileNetV3 \\
        0.6738 & AirNet + DPN107 + GoogLeNet + ShuffleNet \\
        \bottomrule
    \end{tabular}
\end{table}
 
\begin{figure}[!t]
    \centering
    \includegraphics[width=0.95\linewidth]{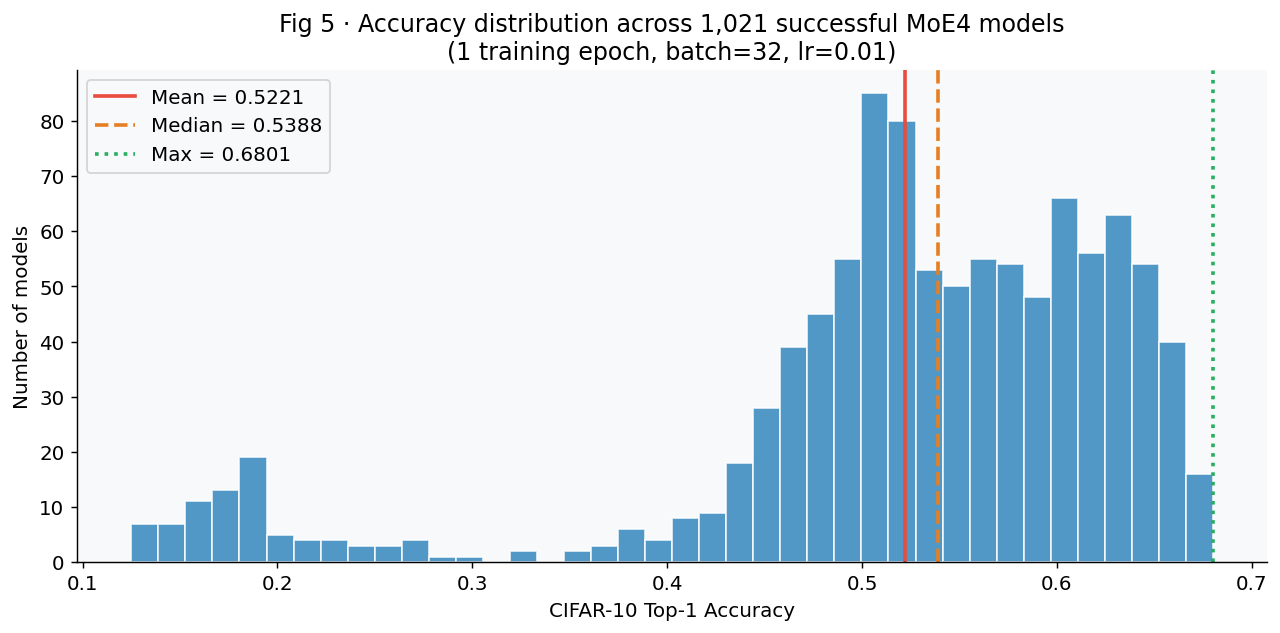}
    \caption{Accuracy distribution across 1,021 successful MoE4 models
    (1 training epoch, batch=32, lr=0.01). Red line: mean (0.5221),
    dashed orange: median (0.5388), dotted green: maximum (0.6801).}
    \label{fig:acc_dist}
\end{figure}

Across the 1,021 successful models, top-1 accuracy ranges from 0.125 to
\textbf{0.6801} with a mean of 0.5221 and median of 0.5388
(Table~\ref{tab:accuracy}, Fig.~\ref{fig:acc_dist}).
The distribution is left-skewed, with a sharp low-accuracy cluster
corresponding to combinations containing MNASNet (mean accuracy 0.218)
and a broad high-accuracy region above 0.55.
The best model --- AirNet + AlexNet + DPN68 + ResNet --- achieves
\textbf{68.0\% top-1 accuracy in a single training epoch}
(Table~\ref{tab:top5}).
All top-5 models share ShuffleNet or MobileNetV3 as a co-expert,
consistent with the family-level analysis below.
 
\subsection{Family-Level Analysis}
 
\begin{table}[!t]
    \caption{Expert family accuracy profile (families with $\geq$40
    appearances in successful models).}
    \label{tab:families}
    \centering
    \fontsize{7.5}{7.5}\selectfont
    \newcolumntype{F}{ >{\raggedright\arraybackslash} m{2.0cm} }
    \newcolumntype{C}{ >{\centering\arraybackslash} m{1.0cm} }
    \newcolumntype{S}{ >{\centering\arraybackslash} m{1.2cm} }
    \newcolumntype{V}{ >{\centering\arraybackslash} m{1.0cm} }
    \begin{tabular}{F C S V}
        \toprule
        \textbf{Family} & \textbf{Mean Acc} & \textbf{Success Rate} &
        \textbf{Apps.} \\
        \midrule
        ShuffleNet      & \textbf{0.632} & 96.7\% & 92  \\
        MobileNetV3     & 0.621          & 96.4\% & 84  \\
        MobileNetV2     & 0.586          & ---    & 72  \\
        InceptionV3-1   & 0.582          & 27.2\% & 320 \\
        ResNet          & 0.578          & 96.8\% & 93  \\
        EfficientNet    & 0.556          & ---    & 81  \\
        DenseNet        & 0.555          & ---    & 79  \\
        SwinTransformer & 0.493          & ---    & 91  \\
        FractalNet      & ---            & 0.7\%  & 3,273 \\
        MNASNet         & 0.218          & 70.1\% & 134 \\
        \bottomrule
    \end{tabular}
\end{table}
 
\begin{figure}[!t]
    \centering
    \includegraphics[width=0.95\linewidth]{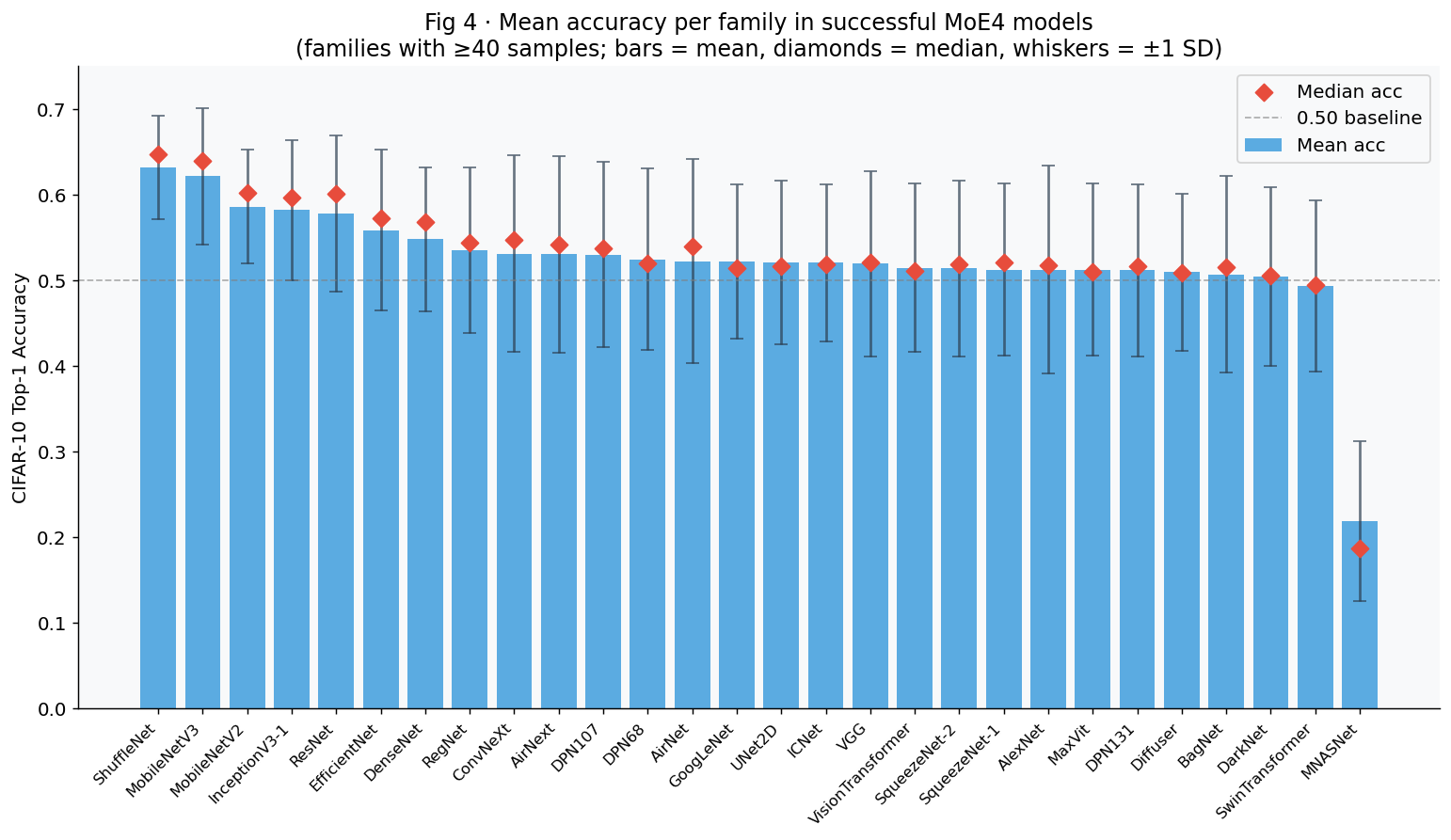}
    \caption{Mean and median accuracy per expert family in successful
    MoE4 models (families with $\geq$40 appearances; bars = mean,
    diamonds = median, whiskers = $\pm$1 SD).}
    \label{fig:family_acc}
\end{figure}

Table~\ref{tab:families} and Fig.~\ref{fig:family_acc} show the
family-level accuracy profile across successful models.
Four key observations emerge:
 
\textbf{ShuffleNet and MobileNetV3 are the strongest expert families},
achieving mean accuracies of 0.632 and 0.621 respectively with
near-perfect success rates ($>$96\%).
Their lightweight architectures fit comfortably within GPU memory even
when combined with larger experts, and their strong feature representations
appear highly complementary to AirNet.
Future campaigns should seed combinations preferentially with these
families.
 
\textbf{FractalNet is a practical dead end.}
Despite being the second most frequently attempted family (3,273
appearances across generated models), FractalNet achieved a success rate
of only \textbf{0.7\%} (23 successful evaluations).
Its recursive architecture causes frequent OOM failures and shape
mismatches at $256{\times}256$ resolution.
Excluding FractalNet from the expert pool would recover a substantial
fraction of wasted compute.
 
\textbf{MNASNet is a consistent accuracy degrader.}
Combinations containing MNASNet produced a mean accuracy of 0.218 ---
above the random baseline of 0.10 for 10-class classification, but far
below the campaign median of 0.539.
This suggests that MNASNet's mobile-optimised architecture, designed for
latency rather than accuracy, does not contribute useful features in a
4-expert ensemble context.
 
\textbf{High success rate does not imply high accuracy.}
SqueezeNet achieves a 96.8\% success rate but only 0.514 mean accuracy,
whereas ResNet achieves the same success rate with 0.578 mean accuracy.
ResNet is therefore the most \emph{efficient} family in this campaign ---
reliable to train and high in accuracy --- making it a strong default
inclusion for future runs.
 
\subsection{General Discussion}
 
All results in this section are valid within a well-defined scope:
AirNet-anchored MoE4 ensembles on CIFAR-10, evaluated after 1 training
epoch.
Claims about family quality (e.g.\ ShuffleNet outperforms SwinTransformer)
cannot be fully decoupled from AirNet's mandatory co-presence until the
coverage bias is corrected in a future campaign.
Nevertheless, the consistency of family rankings across hundreds of
independent combinations --- each with different co-experts --- provides
strong evidence that the observed patterns are genuine and not artefacts
of a specific pairing.
The proposed stratified sampling fix and expert pool pruning
(remove FractalNet and MNASNet) together constitute a concrete,
low-effort improvement that should substantially increase the quality and
coverage of the next campaign.

\section{Conclusion}
\label{sec:conclusion}
This work presented an automated pipeline for large-scale systematic
exploration of heterogeneous 4-Expert Mixture-of-Experts architectures
within the LEMUR neural network dataset ecosystem.
Over a 28-day campaign on an NVIDIA RTX~4090, the pipeline generated and
evaluated \textbf{1,021 successful MoE4 models} across 197 batches,
exceeding the delivery target of 1,000 models and providing a substantial
new collection of expert-combination candidates for the LEMUR database.
 
Three engineering contributions enabled this scale: a deterministic
code-assembly generator that produces syntactically valid, forward-pass
verified MoE4 models from any combination of LEMUR base architecture
families; a multi-stage validation pipeline (syntax check, CPU
forward-pass probe at training resolution, database deduplication) that
eliminated wasted GPU evaluations from broken or duplicate candidates;
and a fault-tolerant campaign orchestrator with persistent state and
SSH-safe resume semantics that sustained 28 days of unattended execution.
 
The campaign's most significant finding is a \textbf{search space
coverage bias} induced by sorted deterministic enumeration: all 1,146
explored combinations were anchored to AirNet, covering only 4.8\% of
the 23,751 possible 4-family quartets and leaving 95.2\% of the search
space entirely unexplored.
This bias is not specific to our pipeline --- it is a general risk in any
large-scale NAS campaign that combines sorted iteration with a
throughput-based stop condition.
The proposed fix, stratified random shuffling of the combination list
before iteration, is a one-line code change that eliminates the bias
entirely.
 
Within the AirNet-anchored scope, family-level analysis across 1,021
models identified \textbf{ShuffleNet} and \textbf{MobileNetV3} as the
highest-yield expert families (mean accuracies 0.632 and 0.621,
respectively, with $>$96\% training success rates), and
\textbf{FractalNet} and \textbf{MNASNet} as families that consistently
waste computation or degrade ensemble quality.
The best single combination --- AirNet + AlexNet + DPN68 + ResNet ---
achieved \textbf{68.0\% Top-1 accuracy in CIFAR-10 in a single training
epoch}, demonstrating that automated heterogeneous expert assembly can
produce competitive models without any manual architecture design.
 
Future work will apply the stratified sampling fix and pruned expert pool
to conduct an unbiased campaign across the full combination space,
extend evaluation beyond single-epoch screening to identify genuinely
high-performing ensembles, and investigate memory-aware pre-filtering
to reduce the dominant CUDA OOM failure rate.

{
    \small
    \bibliographystyle{ieeenat_fullname}
    \bibliography{bibmain}
}

\end{document}